\documentclass[sigconf]{acmart}

\usepackage{amsmath, amsfonts, amssymb, booktabs}
\usepackage{algorithm}
\usepackage{graphicx}
\usepackage{breqn}
\usepackage{listings}
\usepackage{algpseudocode}
\usepackage{multirow}
\usepackage{footnote}

\AtBeginDocument{%
  }

\setcopyright{iw3c2w3}
\copyrightyear{2024}

\acmConference{Preprint}{Working Paper}{Nov 2024}

\title{Towards Objective and Unbiased Decision Assessments with LLM-Enhanced Hierarchical Attention Networks}

\author{
Junhua Liu\textsuperscript{1,2,$*$}, 
Kwan Hui Lim\textsuperscript{2}, 
Roy Ka-Wei Lee\textsuperscript{2}
}
\affiliation{
 \institution{\textsuperscript{1}Forth AI}
 \institution{\textsuperscript{2}Singapore University of Technology and Design}
}

\renewcommand{\shortauthors}{Liu et al.}

\begin{abstract}
How objective and unbiased are we while making decisions? This work investigates cognitive bias identification in high-stake decision making process by human experts, questioning its effectiveness  in real-world settings, such as candidates assessments for university admission. We begin with a statistical analysis assessing correlations among different decision points among in the current process, which discovers discrepancies that imply cognitive bias and inconsistency in decisions. This motivates our exploration of bias-aware AI-augmented workflow that surpass human judgment. We propose BGM-HAN, an enhanced Hierarchical Attention Network with Byte-Pair Encoding, Gated Residual Connections and Multi-Head Attention. Using it as a backbone model, we further propose a Shortlist-Analyse-Recommend (SAR) agentic workflow, which simulate real-world decision-making. In our experiments, both the proposed model and the agentic workflow significantly improves on both human judgment and alternative models, validated with real-world data. Source code is available at: \url{https://github.com/junhua/bgm-han}.
\end{abstract}



\begin{document}
\maketitle

\def\thefootnote{*}\footnotetext{Corresponding author: j@forth.ai}
\def\thefootnote{\arabic{footnote}}

\begin{figure}[t]
    \centering
    \includegraphics[width=\linewidth]{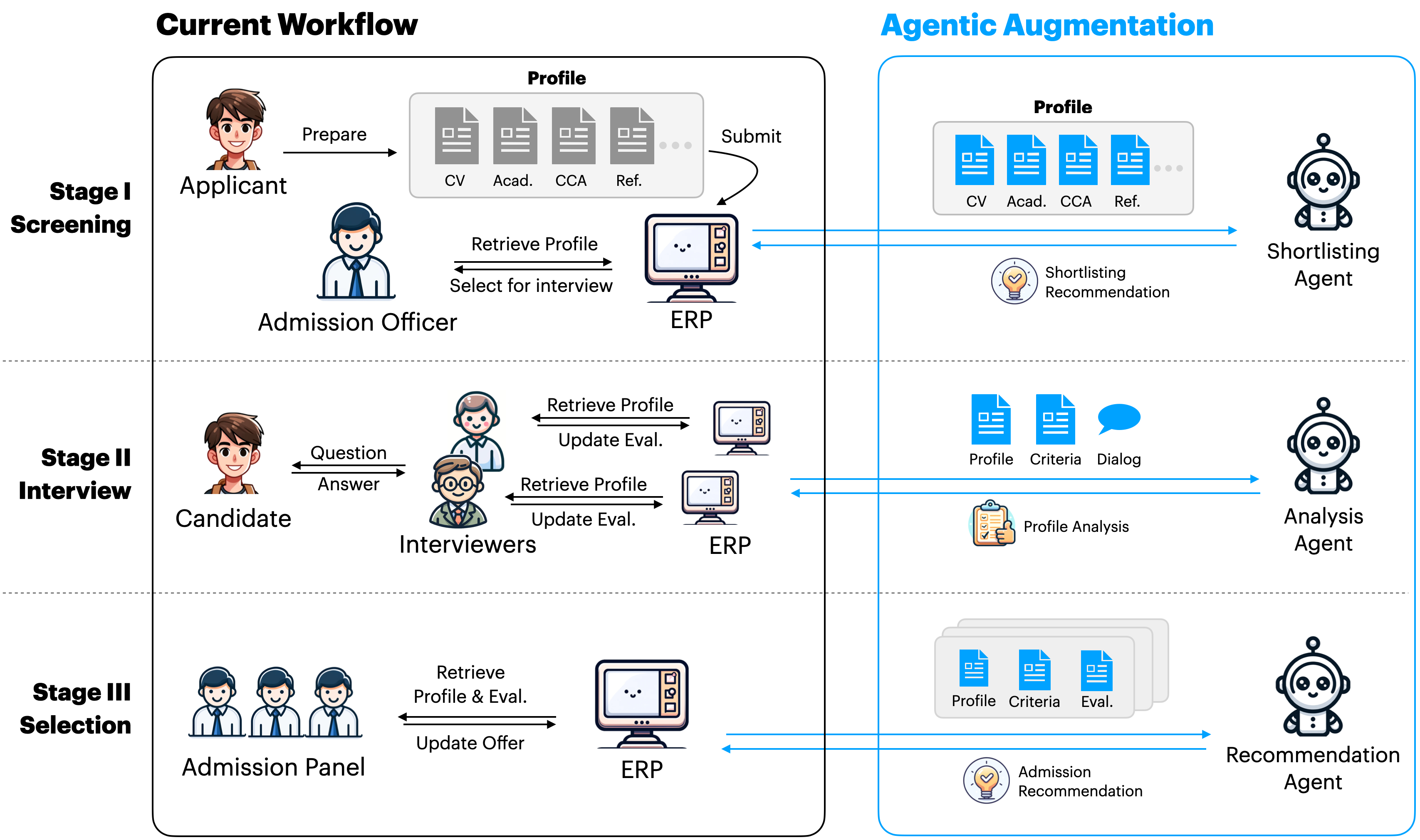}
    \caption{University Admission Decision Process: overview of current workflow and possible agentic augmentation}
    \label{fig:teaser}
    \vspace{-.4cm}
\end{figure}

\section{Introduction}

\begin{table*}[t]
\small
\begin{tabular}{p{1.2cm}p{7cm}p{8cm}}
\toprule
\textbf{Field} & \textbf{Format} & \textbf{Sample Entry} \\
\midrule
GCEA & {[GCEA] School:\{school\}, UAS:\{score\}; Grades:\{subject grade pairs\}} & [GCEA]: School:HCI, UAS:90.0; Grades:H1 PROJECT WORK A, H1 GENERAL PAPER A, H2 ECONOMICS A, H2 CHEMISTRY A, ... \\
\midrule
GCEO & {[GCEO] \{subject-grade pairs\}} & [GCEO] HIGHER CHINESE B3, ENGLISH A1, ELEMENTARY MATHEMATICS A2, ... \\
\midrule
Leadership & {[Leadership] \{activity\}, Level:\{role\}, Year:\{year\}, Category:\{type\}, Participation:\{level\}}. (Multiple entries) & [Leadership] Mind Sports Club, Level:Captain, Year:2023, Category:Sports, Participation:Executive Committee \\
\midrule
PIQ & {[PIQ\{n\}]\{essay text\}}, where n = 1,2,3,4,5 & [PIQ1] My first visit to the university in Primary 5 for a Learning Journey left me with a deep impression... \\
\midrule
OfferType & Offered / Not Offered & Offered \\
\midrule
Analysis & Structured text (profile analysis by LLM) & This candidate demonstrates a strong academic profile, particularly in STEM, with straight As... \\
\bottomrule
\end{tabular}
\caption{Data Format and Sample Entry}
\label{tab:data_format}
\end{table*}

Decision making in high-stake scenarios is often done by human experts leveraging their domain expertise and experience to optimize decision quality~\cite{alur2023auditing}. However, subjectivity and cognitive biases, such as anchoring bias~\cite{haag2024overcoming} and confirmation bias~\cite{echterhoff2024cognitive}, are often difficult to detect and avoid~\cite{kahneman2023cognitive}. Mitigating decision biases is critical to ensure long-term sustainable outcomes and fairness to stakeholders, especially in high-stakes environments~\cite{ghai2022d-bias}.

Recent studies propose artificial intelligence (AI) systems to enhance manual decision processes, such as fairness-aware AI systems guiding decision-makers toward more impartial choices~\cite{yang2024fair}, Explainable AI frameworks that identify potential biases and improve judgment accuracy~\cite{haag2024overcoming}, and  a human-AI collaborative to audit and mitigate social biases~\cite{ghai2022d-bias}. 

Despite the promising recent work in augmenting human expertise and mitigating various biases, the overall advancements are underwhelming due to several reasons. Firstly, the complexity and context-specific nature of cognitive biases make it challenging for AI systems to accurately detect and mitigate~\cite{kahneman2023cognitive}. Secondly, the limited interpretability of AI models can reduce trust in high-stakes settings~\cite{haag2024overcoming}. Lastly, real world scenarios often rely on sensitive or proprietary data. The lack of accessible data impose tremendous challenges in such research activities~\cite{springer2024biasfairness}.

In this regard, this work aims to contribute to \textbf{AI-augmented, bias-aware decision making} in a real world setting, such as university admissions. Current processes involve complex, semi-structured data that requires nuanced assessment across multiple dimensions. Through statistical analysis, we identified non-trivial discrepancies between human evaluations and final outcomes, suggesting inconsistencies and cognitive biases. This motivates our approach, which seeks not only to automate but also to enhance decision consistency and reduce subjective influences through structured AI interventions.

We propose \textbf{BGM-HAN} an enhanced Hierarchical Attention Network with Byte-pair Encoded, Gated Residual Connetions and Multihead Attention, which takes an leverage hierarchical learning approach to better capture and interpret multi-level semi-structured data. Using BGM-HAN as backbone, we introduce a Shortlist-Analyze-Recommend (SAR) agentic workflow to simulate existing human decision processes. 

In our experiments, the proposed models outperform different categories of baseline models. While comparing to current human evaluation, our proposed workflow introduces over 9.6\% improvement in F1-score and accuracy. The promising results uncovers potentials in integrating hierarchical learning with LLM-augmentation to perform automated decision making with implicit fairness and consistency in high-stakes, real-world decision-making environments.

\textbf{Contributions.} In summary, this paper makes the following novel contributions:

Firstly, we propose a statistical approach to identify cognitive bias and inconsistency in a real-world decision making process, i.e., university admission assessments.

Secondly, we propose a hierarchical learning model, \texttt{BGM-HAN}, that effectively represents multi-level semi-structured data and experimentally outperforms baseline models of multiple categories.

Lastly, we propose an agentic workflow, \texttt{$\mathcal{W}_{SAR}$}, that mimics existing human decision process. $\mathcal{W}_{SAR}$ mitigates inconsistency and cognitive bias across different decision makers, and empirically outperforms human evaluation by over 9.6\% in F1-score and accuracy for in decision recommendation.

\section{Problem Formulation}

In the university admissions scenario, the input space consists of student profiles $\mathcal{P} = \{p_1, \ldots, p_n\}$, where each profile $p_i$ comprises four key components: GCE A-Level results ($f_{\text{GCEA}}$), GCE O-Level results ($f_{\text{GCEO}}$), leadership records ($f_{\text{Leadership}}$), and Personal Insight Questions responses ($f_{\text{PIQ}}$). These components present varying processing challenges. Academic records ($f_{\text{GCEA}}, f_{\text{GCEO}}$) contain structured grade information that requires normalization across different subjects and years. Leadership records ($f_{\text{Leadership}}$) present semi-structured text detailing roles, years, categories, and participation levels. PIQ responses ($f_{\text{PIQ}}$) consist of unstructured text requiring sophisticated semantic understanding to evaluate candidates' motivation, challenges, creativity, uniqueness, and institutional fit.

The objective is to develop a mapping $\mathcal{D}: \mathcal{P} \rightarrow \{0, 1\}$ that transforms each profile into an admission decision, where 1 represents an offer and 0 represents a rejection. This mapping must optimize for fairness by eliminating cognitive biases, consistency in treating similar profiles, and interpretability in providing transparent decision rationale.






\begin{figure*}
    \centering
    \includegraphics[width=.99\linewidth]{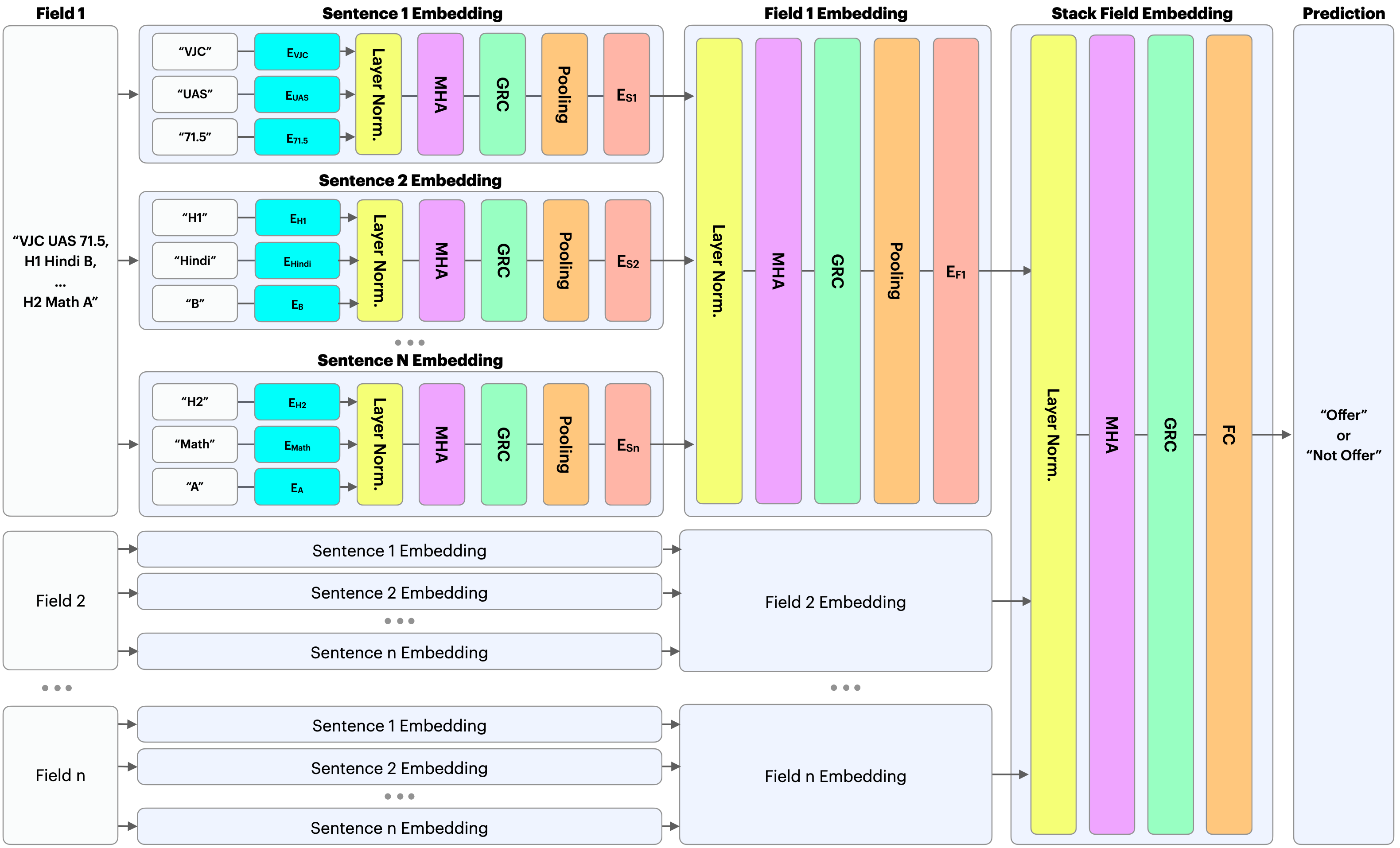}
    \caption{Architecture overview of the proposed BGM-HAN model. The multi-level model learns features from token to sentence to field. At each level, the data will go through layer normalisation, multi-head self-attention (MHA), gated residual connection (GRC), mean pooling to form the higher level embeddings. The embeddings are then concatenated and reshaped into 3D tensors to continue with the next level processing. }
    \label{fig:bgmhan}
\end{figure*}

\begin{figure}[t]
    \centering
    \includegraphics[width=\linewidth]{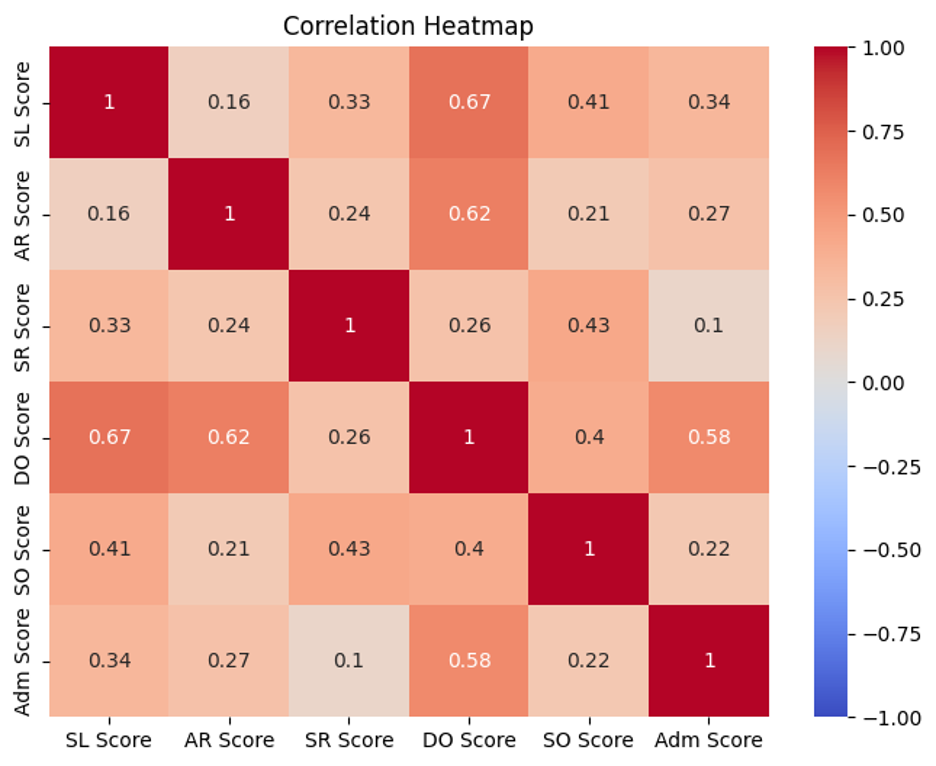}
    \caption{Correlation matrix of different decision points in the admission assessments process.}
    \label{fig:corr}
\end{figure}

\section{Data Analysis}
\label{sec:data}
This section discusses the admission dataset and the statistical analyses performed on the data.

\subsection{Dataset}
Our experiments utilize a corpus that consists of real-world university admissions data, with  student profiles from the 2024 admission cycle. Each profile in our dataset integrates four key components essential for admission decisions: academic records, leadership experiences, personal insight questions (PIQ), and final admission decisions. Table~\ref{tab:data_format} summarizes the data fields, formats and samples.

The academic records encompass both GCE A-Level (GCEA) and O-Level (GCEO) results. GCEA include elements such as high school, University Admission Score (UAS) and detailed subject grades across H1, H2, and H3 levels, as well as GCE O-Level results, providing a comprehensive view of academic achievement.

Leadership experiences are documented in a semi-structured format, capturing roles, positions, years of involvement, and participation levels across various categories such as Sports, Performing Arts, and Community Service. The Personal Insight Questions component consists of five essay responses where candidates articulate their motivation for application, describe significant challenges they've overcome, showcase creative achievements, highlight unique qualities, and explain their institutional fit. Each profile is labeled with a binary admission decision indicating whether an offer was made (1) or not (0).

For profiles receiving offers, we additionally collect detailed analyses generated by experienced evaluators. These analyses provide structured rationale for decisions, highlighting key strengths and potential concerns in each profile. This analytical component serves dual purposes: providing training data for the Analysis Agent ($\mathcal{A}_A$) and establishing a benchmark for evaluating the quality of automated analyses.

\subsection{Decisions Correlation Analysis}
While investigating human decision quality in the current admission process, we first study the correlations among difference decision points in the admission process. The decision points include Shortlisting (SL), Admission Recommendation (AR), Scholarship Recommendation (SR), Degree Offered (DO), Scholarship Offered (SO) and Admission Outcome (Adm). SL provides a priliminary screening process to determine if the students are fit for interview. AR and SR are made based on the findings of structured interviews by the interviewers who are either faculty members or university staffs. DO and SO are the final decision points by a admission committee to decide if the students receive a degree offer and a scholarship offer. The options from each category\footnotemark are ranked into scores and used to calculate the correlation matrix, as shown in Figure~\ref{fig:corr}
\footnotetext{Due to confidentiality, the content of the options are not disclosed.}.

Intuitively, we expect the admission decision (i.e. DO) to adhere to SL and AR, where students who are shortlisted and/or recommended for admission should have a high ratio of receiving degree offers. However, looking at Figure~\ref{fig:corr}, despite the strong correlations between SL and DO (0.67) and between AR and DO (0.62), they are far from perfect correlations of 1.0. 

Therefore, the observation suggests variability in evaluations, which could be influenced by subjectivity and cognitive biases~\cite{phillips2019cognitive}. For instance, interviewers’ subjective judgments during AR and SR could be affected by biases such as anchoring, stereotyping, confirmation bias or the halo effect~\cite{kahneman2023cognitive}, leading to inconsistencies.

\section{Methodology}
\subsection{BGM-HAN}

The architecture overview of BGM-HAN is shown in Figure~\ref{fig:bgmhan}, which later serves as the backbone of the $\mathcal{A}_S$ and $\mathcal{A}_R$ agents (introduced in section~\ref{section:agentic}). We discuss the motivation and implementation of each enhancement component in detail in this section. 

\subsubsection{Base Architecture}
BGM-HAN is designed on a base architecture inspired by Hierarchical Attention Network(HAN)~\cite{yang2016han}, which demonstrated proficiency in capturing the latent information of textual data where its structure embeds additional insights. Candidate profiles consist of multi-level textual information in a semi-structured manner. For example, in academic records, subjects and grades form an entry and multiple entries form a field; similarly in leadership records, Activity, Level, Year and Category form an entry, and zero to many entries form a field (refer to section~\ref{sec:data} and table~\ref{tab:data_format} for more detail). On the other hand, HAN’s dual-level attention mechanisms at both entry and field levels enable the model to focus on the most informative parts of the text across hierarchy. We observe high relevance between neural architecture of HAN and the multi-level semi-structured nature of our data. This capability is particularly crucial for candidates assessments and decision recommendations, where key insights that influence decisions may be dispersed throughout different sections of an applicant’s profile. 

Basing on HAN, we further introduce effective enhancements including widely adopted mechanisms such as byte-pair encoding~\cite{zouhar2023formal}, multi-head attention~\cite{vaswani2017attention} and gated residual connections~\cite{savarese2016learning} to improve the network's ability to capture complex patterns in the text. As a result, BGM-HAN allows each agent to process textual data through multiple hierarchical levels.

\subsubsection{Byte-Pair Encoding and Embedding}
To effectively handle the diverse and variable-length textual data in student profiles, we employ a two-stage tokenization (Algorithm~\ref{alg:bpe}) and embedding (Algorithm~\ref{alg:embedding}) process. 

We choose Byte-Pair Encoding (BPE) as the tokenizer as it shows superior ability in handling out-of-vocabulary issues, which makes it popular among state-of-the-art LLMs, such as the LLaMa3~\cite{dubey2024llama} and GPT-4~\cite{openai2024gpt4technicalreport}. BPE first creates a subword vocabulary of size $V=5000$, then iteratively merges the most frequent pairs of symbols in the data, enabling effective representation of both common and rare words while minimizing the out-of-vocabulary problem.

Using this learned BPE vocabulary, we then transform each text field into a fixed-dimensional tensor through a hierarchical embedding process described in Algorithm~\ref{alg:embedding}. The process maintains the structural hierarchy of the text by operating at sentence and word levels, with dimension constraints $(s,w,d)=(10,50,768)$ for maximum sentences, words per sentence, and embedding dimension. 

Each field embedding $\mathbf{E}_f \in \mathbb{R}^{s \times w \times d}$ is constructed through consistent padding and truncation operations at both word and sentence levels, ensuring uniform tensor dimensions across varying input lengths. This hierarchical representation preserves both local (word-level) and global (sentence-level) semantic information, providing a rich foundation for the subsequent attention mechanisms.

\begin{algorithm}[t]
\caption{Byte-Pair Encoding (BPE)}
\label{alg:bpe}
\begin{algorithmic}[1]
\Require Corpus $\mathcal{C}$ as a sequence of characters, initial vocabulary $\mathcal{V}_0 = \{c : c \in \text{unique characters in } \mathcal{C}\}$, target vocabulary size $N$
\Ensure Final vocabulary $\mathcal{V}$ containing original characters and merged symbols
\State Initialize vocabulary $\mathcal{V} \gets \mathcal{V}_0$
\While{$|\mathcal{V}| < N$}
    \State \textbf{Identify most frequent pair:}
    \For{each consecutive pair of symbols $(a, b)$ in $\mathcal{C}$}
        \State Calculate frequency $f(a, b)$
    \EndFor
    \State Find $(a^*, b^*) = \operatorname*{arg\,max}_{(a, b)} f(a, b)$
    \Comment{$(a^*, b^*)$ is the pair with highest frequency}
    \State \textbf{Merge the pair:}
    \State Define new symbol $s = a^*b^*$
    \State Replace each occurrence of $(a^*, b^*)$ in $\mathcal{C}$ with $s$, forming $\mathcal{C}'$
    \State Update corpus: $\mathcal{C} \gets \mathcal{C}'$
    \State Update vocabulary: $\mathcal{V} \gets \mathcal{V} \cup \{s\}$
\EndWhile
\State \Return Final vocabulary $\mathcal{V}$
\end{algorithmic}
\end{algorithm}

\begin{algorithm}[t]
\caption{Hierarchical Field Embedding}
\label{alg:embedding}
\begin{algorithmic}[1]
\Require Text field $f$, vocabulary size $V$, embedding dimension $d$, maximum sentences $s$, maximum words $w$
\Ensure Field embedding tensor $\mathbf{E}_f \in \mathbb{R}^{s \times w \times d}$
\State Initialize empty sentence embeddings list $\mathcal{S} = []$
\State Split text into sentences: $\{s_1, ..., s_n\} \gets \text{split}(f, \text{delimiter}='.')$

\For{each sentence $s_i$ in $\mathcal{S}_{\text{valid}}[1:s]$}
    \State Apply BPE tokenization: $\text{tokens} \gets \text{BPE}(s_i)$
    \State Convert to tensor: $\mathbf{t} \gets \text{tensor}(\text{tokens})$
    \State Get word embeddings: $\mathbf{W} \gets \text{Embed}(\mathbf{t}) \in \mathbb{R}^{|\text{tokens}| \times d}$
    
    \If{$|\text{tokens}| > w$}  \Comment{Truncate if too long}
        \State $\mathbf{W} \gets \mathbf{W}_{1:w}$
    \ElsIf{$|\text{tokens}| < w$}  \Comment{Pad if too short}
        \State $\mathbf{P} \gets \mathbf{0}_{(w-|\text{tokens}|) \times d}$  \Comment{Create zero padding}
        \State $\mathbf{W} \gets [\mathbf{W}; \mathbf{P}]$  \Comment{Concatenate padding}
    \EndIf
    
    \State Append to sentence list: $\mathcal{S}.\text{append}(\mathbf{W})$
\EndFor

\While{$|\mathcal{S}| < s$}  \Comment{Pad sentence dimension}
    \State $\mathbf{P}_s \gets \mathbf{0}_{w \times d}$  \Comment{Create sentence padding}
    \State $\mathcal{S}.\text{append}(\mathbf{P}_s)$
\EndWhile

\State Stack sentences: $\mathbf{E}_f \gets \text{stack}(\mathcal{S})$  \Comment{Shape: $s \times w \times d$}
\State \Return $\mathbf{E}_f$

\end{algorithmic}
\end{algorithm}

\subsubsection{Multi-Head Attention}

We add multi-head attention~\cite{vaswani2017attention} to capture multiple dependencies and interactions within the text simultaneously, allowing the model to attend to different different positions and capture latent patterns and relationships in the data. 

The multi-head attention mechanism processes inputs $\mathbf{X} \in \mathbb{R}^{l \times d}$:

\[
\text{head}_i = \text{Attention}(\mathbf{X}\mathbf{W}_i^Q,\, \mathbf{X}\mathbf{W}_i^K,\, \mathbf{X}\mathbf{W}_i^V)
\]

with learnable parameters $\mathbf{W}_i^Q, \mathbf{W}_i^K, \mathbf{W}_i^V \in \mathbb{R}^{d \times d_k}$, and:

\[
\text{Attention}(\mathbf{Q}, \mathbf{K}, \mathbf{V}) = \text{softmax}\left(\frac{\mathbf{Q}\mathbf{K}^\top}{\sqrt{d_k}}\right)\mathbf{V}
\]

The attention heads' outputs are concatenated and projected:

\[
\text{MultiHead}(\mathbf{X}) = [\text{head}_1; \ldots; \text{head}_h]\mathbf{W}^O
\]

where $\mathbf{W}^O \in \mathbb{R}^{h d_k \times d}$.

\subsubsection{Gated Residual Network}

The gated residual connection is defined as:

\[
\text{GRN}(\mathbf{X}) = \text{LayerNorm}(\gamma \odot \text{FFN}(\mathbf{X}) + \mathbf{X})
\]

where $\gamma \in \mathbb{R}^d$ is a learnable gate parameter, $\odot$ denotes element-wise multiplication, and $\text{FFN}$ is a feed-forward network:

\[
\text{FFN}(\mathbf{X}) = \text{GELU}(\mathbf{X}\mathbf{W}_1 + \mathbf{b}_1)\mathbf{W}_2 + \mathbf{b}_2
\]

\subsection{Agentic Workflow}
\label{section:agentic}
We propose a \textbf{Shortlist-Analyze-Recommend} agentic workflow $\mathcal{W}_{SAR}$ that combines specialized decision agents to automate and enhance the admissions decision-making process. Specifically, the agents are:

\begin{itemize}
    \item \textbf{Shortlisting Agent ($\mathcal{A}_S$)}: Evaluates each profile to determine if the candidate should be shortlisted for further consideration.
    \item \textbf{Analysis Agent ($\mathcal{A}_A$)}: Generates detailed analyses for shortlisted candidates, providing interpretative insights into the candidate's profile.
    \item \textbf{Recommendation Agent ($\mathcal{A}_R$)}: Integrates both the profile and the analysis to make a final recommendation decision.
\end{itemize}

Each agent $\mathcal{A}_i$ is designed to perform a distinct role within the workflow, leveraging an enhanced Hierarchical Attention Network (HAN) for processing textual data. The enhanced HAN incorporates multi-head attention mechanisms and gated residual connections to improve the representation of textual inputs.

Let $\mathcal{P} = \{p_1, \ldots, p_n\}$ represent a set of student profiles, where each profile $p_i$ consists of four text fields:

\[
p_i = \{f_{\text{GCEA}}, f_{\text{GCEO}}, f_{\text{Leadership}}, f_{\text{PIQ}}\}
\]

The workflow $\mathcal{W}_{SAR}$ operates in three main stages, with each stage orchestrated by a specific agent $\mathcal{A}_i$:

\begin{enumerate}
    \item \textbf{Shortlisting Agent ($\mathcal{A}_S$)}: Evaluates each profile $p_i$ to compute a shortlisting probability $P_s(p_i)$. If $P_s(p_i) > \tau$, the candidate is shortlisted for further consideration.
    \item \textbf{Analysis Agent ($\mathcal{A}_A$)}: For shortlisted candidates, generates a detailed analysis $a_i$ using a large language model.
    \item \textbf{Recommendation Agent ($\mathcal{A}_R$)}: Integrates both the profile $p_i$ and the analysis $a_i$ to compute a final recommendation probability $P_r(p_i)$.
\end{enumerate}

Formally, the agentic workflow can be expressed as:

\[
\mathcal{W}_{SAR}(p_i) = \begin{cases}
\mathcal{A}_R(p_i, a_i), & \text{if } \mathcal{A}_S(p_i) > \tau \\
\text{Not recommended}, & \text{otherwise}
\end{cases}
\]

where $a_i = \mathcal{A}_A(p_i)$.

\subsection{Decision Agents}

\subsubsection{Shortlisting Agent ($\mathcal{A}_S$)}

The Shortlisting Agent $\mathcal{A}_S$ processes each profile field $f$ (such as $f_{\text{GCEA}}$, $f_{\text{GCEO}}$, $f_{\text{Leadership}}$, $f_{\text{PIQ}}$) to compute a shortlisting probability. Each field $f$ is first encoded into a hierarchical attention embedding, denoted as $\mathbf{h}_f$, through the enhanced Hierarchical Attention Network (HAN):

\[
\mathbf{h}_f = \text{BGM-HAN}(f)
\]

where $\mathbf{h}_f \in \mathbb{R}^d$ represents the $d$-dimensional embedding for field $f$, capturing the multi-level structure of each field.

The embeddings are then concatenated and passed into a MLP to calculate the shortlisting probability $P_s(p_i)$:

\[
P_s(p_i) = \mathcal{A}_S(p_i) = \sigma\left(\text{MLP}\left([\mathbf{h}_{\text{GCEA}}; \mathbf{h}_{\text{GCEO}}; \mathbf{h}_{\text{Leadership}}; \mathbf{h}_{\text{PIQ}}]\right)\right)
\]

where $\sigma$ is the sigmoid function.

\subsubsection{Analysis Agent ($\mathcal{A}_A$)}

The Analysis Agent $\mathcal{A}_A$ utilizes the \texttt{Gemini-1.5-Pro-002} large language model to generate detailed analyses for shortlisted candidates during data preparation. Each analysis $a_i$ is generated using a structured prompt with generation parameters $\theta$ (see Appendix~\ref{appendix:prompt}):

\[
a_i = \mathrm{LLM}\left(\mathrm{prompt}(p_i); \, \theta \right)
\]

\subsubsection{Recommendation Agent ($\mathcal{A}_R$)}

For shortlisted candidates, the Recommendation Agent $\mathcal{A}_R$ incorporates both the profile and the analysis to compute the recommendation probability:

\[
P_r(p_i) = \mathcal{A}_R(p_i, a_i) = \sigma\left(\text{MLP}\left([\mathbf{h}_{\text{profile}}; \mathbf{h}_{\text{analysis}}]\right)\right)
\]

where $\mathbf{h}_{\text{profile}}$ is the combined embedding of the profile fields, and $\mathbf{h}_{\text{analysis}}$ is the embedding of the analysis text.

\subsection{Training}

\subsubsection{Loss Function}

The agents are trained using a weighted cross-entropy loss to address class imbalance, ensuring appropriate emphasis on minority classes:

\[
\mathcal{L} = -\sum_{i=1}^N w_{y_i} \left( y_i \log(\hat{y}_i) + (1 - y_i) \log(1 - \hat{y}_i) \right)
\]

where $w_{y_i}$ is the class weight for each sample $i$, defined as:

\[
w_{y_i} = \frac{N}{2 N_{y_i}}
\]

with $N_{y_i}$ representing the number of samples in the class of sample $i$. This weighting approach balances the scale of the loss function to handle class imbalance.

\subsubsection{L2 Regularization}

A weight decay factor is added to the loss function to control overfitting and improve generalization. The modified loss function is expressed as:
\[
\mathcal{L}_{\text{reg}} = \mathcal{L} + \lambda \sum_{i=1}^{N} ||\theta_i||^2
\]
where $\lambda$ is the weight decay parameter and $\theta_i$ are model parameters.

\subsection{Inference}

$\mathcal{W}_{SAR}$ processes each profile $p_i$ at inference-time as follows:

\textbf{Shortlisting}: the candidate is shortlisted if $P_s(p_i) = \mathcal{A}_S(p_i) > \tau$.

\textbf{Analysis}: the Analysis Agent $a_i = \mathcal{A}_A(p_i)$ generates analyses.

\textbf{Recommendation}: the final recommendation probability is computed as $P_r(p_i) = \mathcal{A}_R(p_i, a_i)$.

Admission decision $d_i$ is determined by thresholds $\tau$ and $\delta$:
\[
d_i = \begin{cases}
    1, & \text{if } P_s(p_i) > \tau \text{ and } P_r(p_i) > \delta \\
    0, & \text{otherwise}
\end{cases}
\]
\section{Experiments}

We conducted a series of experiments to evaluate the effectiveness of the Agentic Workflow and its constituent Agents. This section details the experimental setup, including the software environment, hardware configuration, and experiment tracking mechanisms.






\subsection{Training Settings}

\subsubsection{Learning Rate Scheduling}

To maintain efficient convergence, we apply a learning rate scheduler that reduces the learning rate by a factor of $\alpha = 0.1$ when validation accuracy does not improve for $k$ consecutive epochs (patience).

Formally, the learning rate at epoch \( t \) is updated as:
\[
\eta_t = \eta_{t-1} \cdot \alpha \quad \text{if no improvement in last } k \text{ epochs}
\]
The minimum learning rate is constrained to \( \eta_{\text{min}} = 10^{-7} \) to avoid premature convergence.

\subsubsection{Gradient Clipping}

To prevent exploding gradients, especially in deep networks, we apply gradient clipping with a maximum norm of 1.0, ensuring stability during training:
\[
\text{clip}(\nabla \mathcal{L}, \, \text{max\_norm}=1.0)
\]

\subsubsection{Early Stopping}

If no improvement in validation accuracy is observed for a maximum of $p = 10$ epochs, training is terminated early to prevent overfitting.

\subsection{Hyperparameter Optimization}
We performed an extensive grid search to optimize the hyperparameters of the BGM-HAN model. The search space encompassed key architectural and training parameters:

\begin{itemize}
    \item Hidden dimension: [256, 512, \textbf{1024}, 2048]
    \item Number of attention heads: [4, \textbf{8}, 16]
    \item Dropout rate: [0.1, 0.2, 0.3, 0.4, 0.5, \textbf{0.6}, 0.7]
    \item Learning rate: [1e-4, 3e-4, 5e-4, \textbf{1e-5}, 3e-5, 5e-5]
    \item Batch size: [8, 16, \textbf{32}, 64]
\end{itemize}

Each configuration was evaluated using early stopping with a patience of 10 epochs to prevent overfitting, with a maximum of 50 epochs per trial. To assess model performance, we use the validation accuracy as the primary metric for selecting the optimal configuration. Gradient clipping is used with a threshold of 1.0 and utilized the AdamW optimizer with a ReduceLROnPlateau scheduler. The optimal hyperparameters were selected based on the highest achieved validation accuracy while considering model stability and convergence characteristics. The optimial set of hyperparameters for BGM-HAN is bolded in the list.

\subsection{Data Processing}
\subsubsection{Analysis Generation}
the Analysis Agent $\mathcal{A}_A$ generates the analyses $a_i$ for each shortlisted profile using the large language model and a structured prompt (see Appendix~\ref{appendix:prompt}). The generated analyses are stored in the dataset and used as additional input features for the Recommendation Agent $\mathcal{A}_R$.

\subsubsection{Handling Missing Data}
Missing values in text fields are replaced with \textit{NaN} to ensure consistent input dimensions across all samples. This approach avoids introducing biases due to varying input lengths from missing data.

\subsubsection{Dataset Splitting}
The dataset is split into training, validation, and test sets with a ratio of 90-5-5 using stratified sampling to maintain class proportions, ensuring balanced evaluation.

\begin{table}[t]
\scalebox{.8}{
\begin{tabular}{@{}ccl@{}}
\toprule
\textbf{Category} & \textbf{Model} & \textbf{Description and Hyperparameters} \\ \midrule
\multirow{2}{*}{\begin{tabular}[c]{@{}c@{}}Traditional\end{tabular}} & XGBoost & Gradient boosting on BERT embeddings \\ \cmidrule(l){2-3} 
 & TF-IDF & TF-IDF vectorization with logistic regression  \\ \midrule
\multirow{4}{*}{\begin{tabular}[c]{@{}c@{}} \\ Neural \\ Networks\end{tabular}} & MLP &  Multi-Layer Perceptron \\ \cmidrule(l){2-3} 
 & BiLSTM-Indv & BiLSTM with individual features embeddings  \\ \cmidrule(l){2-3} 
 & BiLSTM-Concat & BiLSTM with concatenated features embeddings\\ \cmidrule(l){2-3} 
 & HAN & Hierarchical Attention Network \\ \midrule
\multirow{3}{*}{\begin{tabular}[c]{@{}c@{}} \\ Retrieval \end{tabular}} & FAISS (L2) & k-NN search using L2 distance  \\ \cmidrule(l){2-3} 
 & FAISS-CS & k-NN using cosine similarity \\ \cmidrule(l){2-3} 
 & FAISS-CV & category-specific FAISS-CS with voting \\ \midrule
\multirow{2}{*}{LLM} & GPT-4o & Zero-shot classification \\
 & GPT-4o-RA & Retrieval-augmented 5-shot classification \\ \bottomrule
\end{tabular}}
\caption{Baseline categories and algorithms}
\vspace{-0.6cm}
\end{table}

\subsection{Baseline Models}
\subsubsection{Traditional Machine Learning Baselines} Our traditional baselines include XGBoost, which uses concatenated BERT embeddings, and a TF-IDF with logistic regression model that directly processes raw text. These provide a foundational comparison to neural and retrieval-based methods.

\subsubsection{Neural Network Models} Discriminative neural networks such as sequence models~\cite{hochreiter1997lstm, cho2014gru, liu2022title2vec}, attention-based models~\cite{vaswani2017transformer, yang2016han} and pretrained models~\cite{devlin2019bert, liu2019roberta, lample2019xlm} perform well in many text classification tasks. To benchmark, we evaluate several neural architectures, beginning with an MLP that applies ReLU activation to concatenated BERT embeddings~\cite{devlin2019bert}. Next, we assess two bidirectional LSTM (BiLSTM)~\cite{liu2022title2vec} configurations: one that processes concatenated embeddings and another that treats each text field independently. Lastly, a Hierarchical Attention Network (HAN)~\cite{yang2016han} model enables adaptive weighting of text fields, allowing the model to emphasize relevant portions of the input.

\subsubsection{Retrieval-Based Models} \citet{pmlr-v202-basu23a} showed that by breaking down a underlying learning task into local sub-tasks, retrieval-based models with a low complexity parametric component performance well in classification tasks. Motivated by this work, we include the retrieval-based models as baselines using FAISS by Meta Research~\cite{douze2024faiss}. We implement FAISS-based k-NN models, using both L2 distance and cosine similarity. Additionally, a category-specific k-NN model performs weighted voting across different text fields, tailoring retrieval to each field’s characteristics.

\subsubsection{Large Language Models} Recent LLMs~\cite{openai2024gpt4technicalreport, dubey2024llama, anthropic2024claude3} showed superior performance in general natural language understanding and generation tasks. We intend to investigate pretrained LLMs' ability to perform zero-shot and few-shot classification without finetuning. Specifically, we choose the best LLM i.e. GPT-4o~\cite{openai2024gpt4technicalreport} in two settings: zero-shot classification and a Retrieval-Augmented Generation (RAG) approach. The former investigates LLM's classification ability by implicit knowledge, while the latter examines the effect of in-context learning on improving classification performance. 

\subsubsection{Hyperparameters and evaluation metrics} Each baseline model processes fields including high school grades, middle school grades, leadership records, and self-assessments. The BERT embeddings are generated using \texttt{bert-base-uncased} model with a maximum sequence length of 512 tokens. For neural models, we use the Adam optimizer with a learning rate $2\times10^{-5}$ and train for up to 100 epochs with early stopping triggered by a moderate patience of 10 epochs. Model performance is evaluated using accuracy, precision, recall, F1-score, and confusion matrices, providing a comprehensive assessment of the agents' predictive capabilities and ensuring both high precision and recall.

\begin{table}[t]
\centering
\scalebox{1}{
\begin{tabular}{l|c|c|c|c}

\toprule
\textbf{Model}                  & \textbf{Precision} & \textbf{Recall} & \textbf{F1} & \textbf{Accuracy} \\ \midrule
\multicolumn{5}{c}{\textbf{Human Evaluation}} \\ \midrule
Shortlisting                         & 0.8464                         & 0.8418                      & 0.8156                         & 0.8155            \\ \midrule
Interview-Rec           & 0.8011                         & 0.8193                      & 0.8321                         & 0.8087            \\ \midrule
\multicolumn{5}{c}{\textbf{Traditional Machine Learning Models}} \\ \midrule
XGBoost                         & 0.7902                         & 0.7859                      & 0.7878                         & 0.7931            \\ \midrule
TF-IDF           & 0.6938                         & 0.6527                      & 0.6488                         & 0.6839            \\ \midrule

\multicolumn{5}{c}{\textbf{Neural Network Models}} \\ \midrule
MLP                             & 0.7967                         & 0.7990                      & 0.7911                         & 0.7989            \\ \midrule
HAN                             & 0.7716                         & 0.7707                      & 0.7711                         & 0.7759            \\ \midrule

BiLSTM-Indv          & 0.7963                         & 0.7612                      & 0.7667                         & 0.7816            \\ \midrule
BiLSTM-Concat                   & 0.8291                         & 0.8178                      & 0.8176                         & 0.8276            \\ \midrule

\multicolumn{5}{c}{\textbf{Retrieval-Based Models}} \\ \midrule
FAISS-L2             & 0.6886                         & 0.6707                      & 0.6897                         & 0.6897            \\ \midrule
FAISS-CS       & 0.6659                         & 0.6724                      & 0.6654                         & 0.6713            \\ \midrule
FAISS-CV      & 0.7796                         & 0.7511                      & 0.7558                         & 0.7640            \\ \midrule

\multicolumn{5}{c}{\textbf{Large Language Models}} \\ \midrule
GPT-4o                           & 0.5579                         & 0.5114                      & 0.4111                         & 0.5600            \\ \midrule
GPT-4o-RA                       & 0.7347                         & 0.7365                      & 0.7352                         & 0.7371            \\ \midrule

\multicolumn{5}{c}{\textbf{Proposed Models}} \\ \midrule
BGM-HAN              & \underline{0.8622}                         & \underline{0.8405}                      & \underline{0.8453}                         & \underline{0.8506}            \\ \midrule
BGM-HAN-$\mathcal{W}_{SAR}$          & \textbf{0.8995}                         & \textbf{0.8918}                      & \textbf{0.8945}                         & \textbf{0.8966}            \\ \bottomrule
\end{tabular}}
\caption{Summary of Experimental results. The highest values are in bold whereas the second-highest are underlined.}
\label{table:experiment}
\vspace{-.6cm}
\end{table}

\subsection{Experimental Results}
Table~\ref{table:experiment} summarises the experimental results across proposed models, human evaluation, and different categories of baseline models. We discuss our observations and interpretation as follows:

\subsubsection{Proposed Models} BGM-HAN and BGM-HAN-$\mathcal{W}{SAR}$ demonstrate effectiveness in representing hierarchical text features and show the best performance across all metrics. BGM-HAN achieved a macro-averaged F1-score of 0.8453 and accuracy of 0.8506, outperforming all baseline models and approaching human evaluation benchmarks. Enhanced by the proposed agentic workflow,  BGM-HAN-$\mathcal{W}{SAR}$ displays the best results among all, achieving an F1-score of 0.8945 and accuracy of 0.8966. 

\subsubsection{Discriminative Classification} We observe that both traditional discriminative models, such as XGBoost, and neural networks discriminative models, such as BiLSTM, can perform well in the decision assessment tasks. Specifically, XGBoost achieves an F1-score of 0.7878 and accuracy of 0.7931, whereas BiLSTM-Concat performs 0.8176 in F1-score and 0.8276 in accuracy, despite its simple and small structure relatively to LLMs.

\subsubsection{LLMs for Classification} The performance of GPT-4o in a zero-shot is poor, with an F1-score of 0.4111 and accuracy of 0.5600. The GPT-4o with retrieval augmentation (RA) shows significant improvements, achieving an F1-score of 0.7352 and accuracy of 0.7371. This observation suggests that provding relevant context through RA does effectively enhance the LLM's classification ability. However, both models still under-perform as compared to other discriminative approaches. We believe that without domain-specific fine-tuning, LLMs could not perform effective classification by its pre-trained knowledge. 

\subsubsection{Retrieval Approach} Although LLM with RA shows significant improvement to LLM with zero-shot inference, the retrieval-based models perform poorly as compared to all other categories, showing that retrieval alone could not perform well in the classification tasks.

\begin{table}[t]
\centering
\scalebox{.9}{
\begin{tabular}{|l|c|c|}
\hline
\textbf{Backbone}                & \textbf{F1-Score} & \textbf{Accuracy} \\ \hline
\multicolumn{3}{|c|}{\textbf{Traditional Classifier}} \\ \hline
XGBoost                                & 0.7857                         & 0.8076            \\ \hline
TF-IDF        & 0.6291                         & 0.6545            \\ \hline

\multicolumn{3}{|c|}{\textbf{Neural Networks}} \\ \hline
MLP           & 0.7911                         & 0.7989            \\ \hline
HAN   & 0.5833                         & 0.6379            \\ \hline
BiLSTM-Indv   & 0.7667                         & 0.7816            \\ \hline
BiLSTM-Concat           & 0.7637                         & 0.7718            \\ \hline

\multicolumn{3}{|c|}{\textbf{Retrieval Models}} \\ \hline
FAISS-L2                    & 0.6393                         & 0.6430            \\ \hline
FAISS-CS              & 0.5997                         & 0.6246            \\ \hline
FAISS-CV                & 0.7508                         & 0.7606            \\ \hline

\multicolumn{3}{|c|}{\textbf{Proposed Model}} \\ \hline
BGM-HAN                                & \textbf{0.8453}                         & \textbf{0.8506}            \\ \hline
\end{tabular}}
\caption{Ablation Study of Different Backbone Models}
\label{table:ablation-backbone}
\vspace{-.9cm}
\end{table}

\subsection{Ablation Study}

\subsubsection{Component-wise ablation.}
From table~\ref{table:experiment}, we observe that HAN without enhancement has an F1-score and accuracy of 0.7711 and 0.7759 respectively, while our proposed enhancement contributes to 9.6\% improvement, bringing the F1-score and accuracy to 0.8453 and 0.8506. Specifically, we observe approximately 1.8\% improvement by introducing BPE tokenization, 5.2\% by multihead self-attention, and 2.6\% improvement by gated residual connections.

\subsubsection{Backbone models ablation.} 
We conduct ablation study on using other baseline models for $\mathcal{A}_S$ and $\mathcal{A}_R$. The results are summarized in Table~\ref{table:ablation-backbone}. We observe that all other approaches, such as traditional classifiers, neural models and retrieval-based models, perform much poorly than the proposed BGM-HAN. As BGM-HAN's performance, i.e., an F1-score of 0.8453 and accuracy of 0.8506, outperforms all alternative backbones by a large margin, showing BGM-HAN as the optimal backbone for this task.

\subsubsection{LLMs ablation.} 
The proposed agentic workflow $\mathcal{W}_{SAR}$ uses an LLM to generate analysis of the report. We created analyses for all profiles using base models of \texttt{Gemini-1.5-pro}, \texttt{Gemini-1.0-pro}, \texttt{GPT-4o} and \texttt{gpt-4-turbo}. All versions contribute to around 0.049 to 0.051 improvement in accuracy. We chose \texttt{Gemini-1.5-pro} as base LLM model as its impact is marginally better than the rest (See Appendix~\ref{appendix:setting} for model settings.

\section{Related Work}

\textbf{Classification for decision making.} High-stakes decision making is often done by human experts~\cite{alur2023auditing}. \citet{yang2016hierarchical} introduced Hierarchical Attention Network (HAN for document classification, effectively utilizing word and sentence-level attention mechanisms to enhance performance . Building upon this, \citet{ribeiro2020pruning} proposed enhancements to HANs through pruning and Sparsemax to improve interpretability and efficiency in text classification tasks. Various neural networks, such as sequence models~\cite{hochreiter1997lstm, cho2014gru, liu2022title2vec}, attention-based models~\cite{vaswani2017transformer, yang2016han} and pretrained models~\cite{devlin2019bert, liu2019roberta, lample2019xlm}, showed promising classification ability. Recent LLMs show superior performance in general tasks ~\cite{openai2024gpt4technicalreport, dubey2024llama, anthropic2024claude3}, which could be further enhanced by retrieval augmentation\citet{pmlr-v202-basu23a}. 

\textbf{Decision bias.}  The notion of decision-making bias has been critically examined by \cite{phillips-wren2019cognitive}, who analyzed cognitive biases and their impact on decision-making consistency, highlighting the need for unbiased automated systems in high-stakes environments. Algorithmic bias was identified in the predictions in criminal justice~\cite{angwin2016machine,dressel2018accuracy}. Several recent work explored AI-based and data-driven methods to mitigate human biases in decision-making processes. \citet{yang2024fair} propose a fairness-aware AI system to guide individuals toward unbiased decision-making. \citet{haag2024overcoming} proposed and investigated the use explainable AI (XAI) to reduce anchoring bias in consumer decisions. \citet{echterhoff2024cognitive} introduce BiasBuster, a framework for identifying and addressing cognitive biases in large language models, with applications in high-stakes contexts. \citet{ghai2022d-bias} present D-BIAS, a human-in-the-loop tool that applies causal analysis to audit and mitigate social biases, emphasizing the importance of interactive AI in tackling algorithmic bias effectively. 
\section{Impact Statement}
We observe a lack of investigation and discussion bias and fairness in the literature of automated decision making, as compared to performance optimization in general. Our work highlights novelty in a bias-aware GenAI application which yields significant social impact in promoting responsible and fair use of AI.

Particularly, university admissions processes present tremendous challenges given the exhaustive shortlisting, interview and offer decisions for thousands of candidates each year. Our work investigates the possibility of automating such processes in part of in full, while maintaining or even enhancing decision quality.

To ensure practicality and usability of the outcome, the authors work closely with the university admission office throughout the project. All analysis and predictions are based on actual data with anonymization to remove personal identifiable information while retaining its features.

This work is being piloted with the university admission team and scheduled to go in production by the first quarter of 2025, for assessing the upcoming applications. 

The proposed models can be extended to other high-stake scenarios, where decision quality are critical. Examples include assessing job applicants for Human Resource departments, vendor selection in Procurement, and loan approval in financial institutions.
\section{Conclusion}

This study addresses the significant challenge of enhancing objectivity and consistency in high-stakes decision-making processes, such as university admissions, where biases and subjectivity can negatively impact decision outcomes. We proposed the Byte-Pair Encoded, Gated Multi-head Hierarchical Attention Network (BGM-HAN) model, which leverages a hierarchical learning approach to effectively capture and interpret multi-level semi-structured data. Through an extensive data analysis of existing admission decisions, we identified correlations and discrepancies in human evaluative processes, motivating the development of a more consistent, AI-augmented approach to mitigate biases and subjectivity.

The proposed Shortlist-Analyse-Recommend (SAR) agentic workflow combines specialized agents to mimic real-world decision processes while leveraging BGM-HAN as the backbone. Experimentally, the proposed workflow demonstrates superior performance over baseline models, showing over 9.6\% improvement in both F1-score and accuracy compared to human evaluators. This agentic approach aligns with related works advocating for fairness-aware AI and transparency in automated decision systems, further demonstrating how explainable AI frameworks can be integrated to address potential biases in high-stakes decisions. 

Future work may extend to broader applications where decision quality and bias mitigation are critical, such as in HR assessments, financial loan approvals, and vendor selection processes.



\bibliographystyle{ACM-Reference-Format}
\bibliography{custom}

\appendix
\section{Prompt and Model Configurations}
\label{appendix:prompt}

\subsection{Prompt}
The following prompt was used to generate the analysis for each candidate:

\begin{quote}
\footnotesize
You are an experienced university admission officer. Your task is to analyze and summarise this candidate's profile comprehensively across multiple aspects:

\begin{enumerate}
    \item \textbf{Academic Strength Assessment:}
    \begin{itemize}
        \item GCEA Results: \{GCEA Results\}
        \item GCEO Results: \{GCEO Results\}
        \item Focus on performance in STEM subjects
        \item Note any special academic achievements (H3, merit awards, etc.)
    \end{itemize}
    \item \textbf{Technical Aptitude:}
    \begin{itemize}
        \item Evaluate demonstrated interest and capability in STEM
        \item Look for project work, independent learning, or technical activities
        \item Consider any innovative or creative technical solutions mentioned
    \end{itemize}
    \item \textbf{Leadership \& Soft Skills:}
    \{Leadership Experience\}
    \begin{itemize}
        \item Analyze leadership roles and responsibilities
        \item Evaluate team collaboration and project management experience
        \item Consider diversity of leadership experiences
    \end{itemize}
    \item \textbf{Personal Insight Questions Analysis:}
    \{Personal Insight Questions\}
    \begin{itemize}
        \item Assess motivation and alignment with engineering
        \item Evaluate cultural fit with institutional core values: Leadership, Integrity, Passion, Collaboration, Creativity
        \item Look for evidence of red flags or negative traits against institutional core values
    \end{itemize}
\end{enumerate}

Based on these inputs, provide a balanced 150-200 word analysis with precise language, that:
\begin{enumerate}
    \item Highlights key strengths that make them suitable for engineering
    \item Identifies any potential areas of concern
    \item Evaluates their overall fit for an engineering program
    \item Comments on their potential to contribute to a collaborative learning environment
\end{enumerate}

Focus on specific evidence from their profile rather than general statements.
\normalsize
\end{quote}

In the actual prompt, placeholders such as \{\texttt{GCEA Results}\}, \{\texttt{GCEO Results}\}, \{\texttt{Leadership Experience}\}, and \{\texttt{Personal Insight Questions}\} are replaced with the candidate's actual data.

\subsection{Model Configurations}
\label{appendix:setting}
We used the \texttt{gemini-1.5-pro-002} model from Google Generative AI with the following configuration settings:

\begin{itemize}
    \item \textbf{Temperature}: 0.3 (lower temperature for more focused and consistent output)
    \item \textbf{Top-p}: 0.8 (moderate top-p for balanced diversity)
    \item \textbf{Top-k}: 40 (standard top-k for good quality)
    \item \textbf{Max output tokens}: 1024 (sufficient for detailed analysis)
\end{itemize}

The safety settings were configured as shown in Table~\ref{tab:safety_settings} to ensure compliance with content guidelines. This configuration aims to produce detailed and focused analyses while ensuring compliance with safety standards.

\begin{table}[h]
\centering
\small
\begin{tabular}{ll}
\toprule
\textbf{Category} & \textbf{Threshold} \\
\midrule
Harassment & \texttt{BLOCK\_MEDIUM\_AND\_ABOVE} \\
Hate Speech & \texttt{BLOCK\_MEDIUM\_AND\_ABOVE} \\
Sexually Explicit & \texttt{BLOCK\_MEDIUM\_AND\_ABOVE} \\
Dangerous Content & \texttt{BLOCK\_MEDIUM\_AND\_ABOVE} \\
\bottomrule
\end{tabular}
\caption{Safety settings for the generative model.}
\label{tab:safety_settings}
\end{table}

\end{document}